\documentclass[letterpaper]{article} 
\usepackage[submission]{aaai25}  
\usepackage{times}  
\usepackage{helvet}  
\usepackage{courier}  
\usepackage[hyphens]{url}  
\usepackage{graphicx} 
\usepackage{multirow,amssymb,amsmath,pifont}
\usepackage{natbib}  
\usepackage{caption} 
\usepackage{algorithm}
\usepackage{algorithmic}

\usepackage{newfloat}
\usepackage{listings}
\DeclareCaptionStyle{ruled}{labelfont=normalfont,labelsep=colon,strut=off} 
\floatstyle{ruled}
\newfloat{listing}{tb}{lst}{}
\floatname{listing}{Listing}
\title{Stealthy and Robust Backdoor Attack against 3D Point Clouds through Additional Point Features}
\author{
	Xiaoyang Ning\textsuperscript{\rm 1},
	Qing Xie\textsuperscript{\rm 12},
	Jinyu Xu\textsuperscript{\rm 1},
	Wenbo Jiang\textsuperscript{\rm *3},
	Jiachen Li\textsuperscript{\rm 1},
	Yanchun Ma\textsuperscript{\rm 4}
}
\affiliations{
    \textsuperscript{\rm 1}School of Computer Science and Artificial Intelligence, Wuhan University of Technology\\

    \textsuperscript{\rm 2}Engineering Research Center of Intelligent Service Technology for Digital Publishing, Ministry of Education\\
	\textsuperscript{\rm 3}School of Computer Science and Engineering, University of Electronic Science and Technology of China\\
	\textsuperscript{\rm 4}Wuhan Vocational College of Software and Engineering\\
	\{nxyandwill, felixxq, jinyxu\}@whut.edu.cn, wenbo\_jiang@uestc.edu.cn, lijiachen@whut.edu.cn, mayanchun@whvcse.edu.cn
}

\usepackage{bibentry}

\begin{document}

\maketitle

\begin{abstract}
Recently, 3D backdoor attacks have posed a substantial threat to 3D Deep Neural Networks (3D DNNs) designed for 3D point clouds, 
which are extensively deployed in various security-critical applications.
Although the existing 3D backdoor attacks achieved high attack performance,
they remain vulnerable to preprocessing-based defenses (e.g., outlier removal and rotation augmentation) and are prone to detection by human inspection. 
In pursuit of a more challenging-to-defend and stealthy 3D backdoor attack, this paper introduces the \textbf{S}tealthy and \textbf{R}obust \textbf{B}ackdoor \textbf{A}ttack (SRBA), 
which ensures robustness and stealthiness through intentional design considerations.
The key insight of our attack involves applying a uniform shift to the additional point features of point clouds (e.g., reflection intensity) widely utilized as part of inputs 
for 3D DNNs as the trigger. Without altering the geometric information of the point clouds, 
our attack ensures visual consistency between poisoned and benign samples, and
demonstrate robustness against preprocessing-based defenses. In addition, to automate our attack, we employ Bayesian Optimization (BO) to identify the suitable trigger. 
Extensive experiments suggest that SRBA achieves an attack success rate (ASR) exceeding $\mathbf{94\%}$ in all cases, 
and significantly outperforms previous SOTA methods when multiple preprocessing operations are applied during training.
\end{abstract}

\section{Introduction}
Recently, 3D point clouds have been widely applied in various security-critical applications, including autonomous driving, 
robot industry and augmented reality. 
However, 3D backdoor attacks have demonstrated severe and insidious threats to  3D Deep Neural Networks (3D DNNs) specifically designed for point clouds. In typical 3D backdoor attacks, 
attackers embed malicious triggers into some point clouds and assign them  attacker-specified target labels, creating poisoned point cloud datasets. 
These poisoned datasets are then made publicly available. Once unsuspecting users train their 3D DNNs using these poisoned datasets  from unreliable sources, the hidden 
and malicious backdoor is implanted into their 3D DNNs. While backdoored 3D DNNs behave normally with benign point clouds, they activate predetermined
malicious behaviors upon encountering the poisoned point clouds during inference. The insidious yet impactful nature of 3D backdoor attacks poses a significant threat in safety-critical applications.
\begin{figure}[!tbp]
    \centering
    \includegraphics[width=0.45\textwidth]{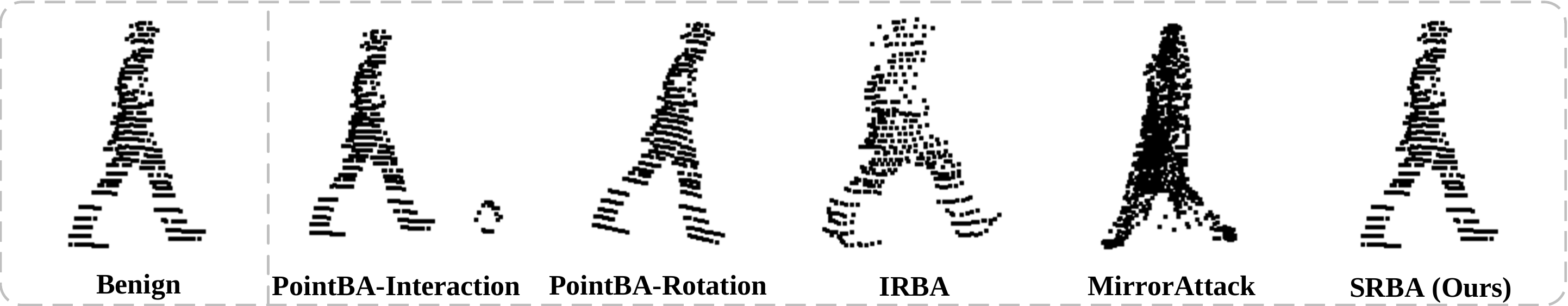}
    \caption{Visual comparison between the benign point cloud and  poisoned point clouds from different 3D backdoor attacks, 
	including  PointBA-Interaction attack~\cite{li2021pointba,xiang2021backdoor}, PointBA-Rotation attack ~\cite{li2021pointba,fan2022careful},
	IRBA~\cite{gao2023imperceptible}, MirrorAttack~\cite{bian2024mirrorattack} and our SRBA.}
	\label{fig:compare}
\end{figure}
\begin{figure*}[th]
	\centering
    \includegraphics[width=0.9\textwidth]{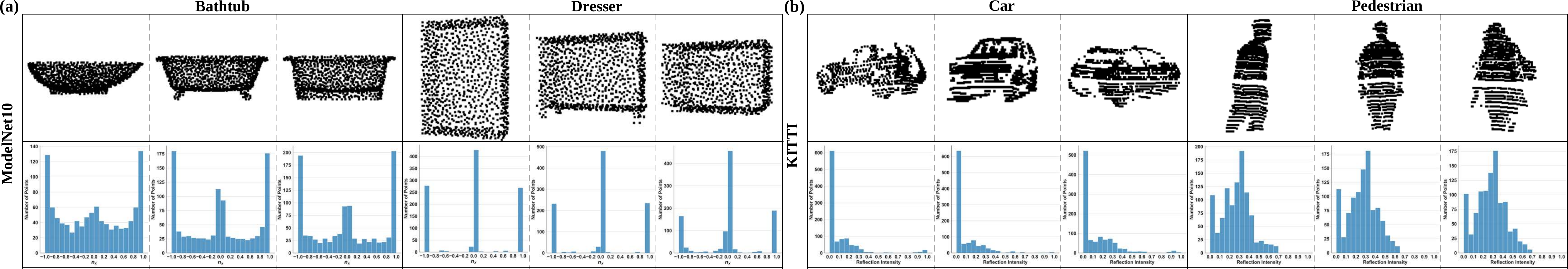}
    \caption{The distributions of additional point features for different categories of point clouds from the ModelNet10 and KITTI datasets. (a) The distributions of x-component of the face normal vector, $n_{\mathrm{x}}$, (b) the distributions of reflection intensity.}
    \label{fig:frequency}
\end{figure*}

Despite extensive research conducted on backdoor attacks, the majority of the existing studies have primarily concentrated on
image domain~\shortcite{gu2017badnets,li2020invisible,zhong2020backdoor,hammoud2021check,wang2021backdoor,jiang2023color}. 
Conversely, research on 3D backdoor attacks remains relatively limited  and suffers from different deficiencies.
Specifically, these 3D backdoor attacks exhibit varying degrees of performance degradation when common preprocessing operations are applied during training. 
Furthermore, their poisoned point clouds are easily distinguishable from their benign versions, making them detectable even through human inspection, as shown in Fig.~\ref{fig:compare}. 
Therefore, \textit{this paper aims to design a stealthy and robust backdoor attack against 3D point clouds (SRBA)}.

We further investigate the underlying reasons for the deficiencies observed in the existing 3D backdoor attacks.
Specifically, point clouds are sets of points 
where each point is a vector consisting of spatial position features $\left( x,y,z \right)$  
and additional point features such as face normal vector and reflection intensity.
Current 3D backdoor attacks typically alter the spatial position features of point clouds through specific operations, such as adding points or applying transformations, 
to embed triggers.
However, these attacks depend on spatial position-based triggers, which are frequently disrupted by preprocessing operations, leading to degraded attack performance.
In detail, PointBA-I~\cite{li2021pointba} and PCBA~\cite{xiang2021backdoor}
embed the ball trigger by adding additional points to the original point clouds, 
yet their triggers are easily removed by statistical outlier removal (SOR)~\cite{zhou2019dup}, a common preprocessing technique. 
Similarly, PointBA-O~\cite{li2021pointba} and NRBdoor~\cite{fan2022careful} introduce  the rotation trigger implemented by rotating benign point clouds,
which are susceptible to disruption by random rotation augmentation. Furthermore, although IRBA~\cite{gao2023imperceptible} and MirrorAttack~\cite{bian2024mirrorattack}
employ more complex ways to embed triggers,
their triggers are also prone to disruption when subjected to multiple preprocessing operations simultaneously.
Moreover, since the geometric information and visual appearance of a point cloud are heavily influenced by the spatial position features of its points, 
the aforementioned attacks, which modify these features to implant triggers,
failed to maintain visual consistency between benign and poisoned point clouds, particularly when encountering more complex point clouds.

To address the aforementioned issues, we consider constructing poisoned samples via the additional point features. 
These features are extensively utilized by 3D DNNs in point cloud classification~\shortcite{qi2017pointnet,zhao2021point,zhao2021point}, 
and 3D object detection~\shortcite{zhou2018voxelnet,yin2021center,yang20203dssd}.
Specifically, we observe that the distributions of additional point feature exhibit similarities within the same classes and differences between different classes, 
as shown in Fig~\ref{fig:frequency}. 
Although 3D DNNs  primarily learn and rely on the geometric information of point clouds for their predictions,
we argue that these models also capture distribution patterns of  additional point features of point clouds in distinct classes,
as potential class-specific information provided by these patterns is beneficial for their predictions.
Consequently, their predictions will be influenced by perturbations in the distributions of additional point feature.
Therefore, the differences generated by perturbing these distributions of point clouds can serve as the trigger. 
Additionally, color backdoor~\cite{jiang2023color}  employed a uniform shift for all pixels to generate poisoned images with the color style distribution
extremely different from that of benign images. It supports ours viewpoint from the perspective of color
style distribution and provides insight into how to generate the difference of additional point feature distributions between poisoned and benign point clouds.

Based on the above analysis, SRBA employs a uniform shift on the additional point features of partial points in point clouds to generate poisoned samples.
Since SRBA embeds its trigger within the additional point features of  point clouds,
it exhibits robust resistance to preprocessing operations targeting the spatial position features. 
Furthermore, our attack maintains visual consistency between benign
and poisoned samples by avoiding modifications to the spatial position features. 
Besides, to automatically search the uniform shift under the practical black-box setting\footnote{The attacker is assumed to have no prior knowledge of the victim model}, 
we then adopt Bayesian Optimization (BO)~\cite{shahriari2015taking}, an effective gradient-free optimization algorithm,
to avoid the need for knowledge of the victim model's architecture and parameters.
Specifically, we use the backdoor loss of a pretraining model 
to evaluate the  effectiveness of one trigger, 
and apply the L1 norm of one trigger as a penalty to constrain its magnitude.

Extensive experiments verify the feasibility of SRBA and demonstrate its high attack performance while maintaining stealthiness.
Additionally, defense experiments underscore the resilience of SRBA against various defensive measures. 
Our main contributions can be summarized as follows:
\begin{itemize}
	\item We are the first to demonstrate the feasibility of embedding backdoor trigger via additional point features of point clouds.
	\item We propose a poisoning-based backdoor attack against 3D point clouds, named SRBA, which exhibits both stealthiness and robustness.
	\item We conduct extensive experiments demonstrating the excellent performance of SRBA  in terms of attack efficacy and resistance against diverse defenses. When compared to previous methods,
 SRBA achieves SOTA performance in bypassing common preprocessing-based defenses.
\end{itemize}

\section{Related Works}
\subsection{Backdoor Attacks against 3D Point Clouds}
PointBA-I~\cite{li2021pointba} and PCBA~\cite{xiang2021backdoor} adopted additional points with a spherical shape as the ball trigger. 
PointBA-O~\cite{li2021pointba} achieved a rotational trigger via 
rotating  point clouds. NRBdoor~\cite{fan2022careful} verified the feasibility of the rotation trigger on 3D meshes. However, the ball trigger and rotation trigger 
could be disrupted by  SOR and random rotation augmentation, respectively. To improve the aforementioned flaws, 
Gao et al. introduced the IRBA~\cite{gao2023imperceptible}, which achieved enhanced robustness to multifarious preprocessing techniques by the weighted local transformation~\cite{kim2021point}. 
However, IRBA faces challenges in maintaining imperceptibility when dealing with  more complex point clouds, and sustain high attack performance under multiple preprocessing operations. 
Similarly, MirrorAttack~\cite{bian2024mirrorattack}, employing a lightweight point cloud reconstruction network~\cite{yang2018foldingnet} to generate poisoned point clouds, 
encounters the same challenges as IRBA.
Besides, IBAPC~\cite{fan2024invisible} and 3TPS~\cite{feng2024stealthy} 
require extensive control over both training data and the training process of 3D DNNs, 
making them less practical compared to our method.
\subsection{Backdoor Defenses against 3D Point Clouds}
Preprocessing operations, such as random rotations, effectively defend against 3D backdoor attacks by disrupting the embedded triggers. 
Beyond preprocessing operations, 
several methods leverage the distinct characteristics of backdoor trigger to detect poisoned samples. 
For example, STRIP~\cite{gao2019strip} and PointCRT~\cite{hu2023pointcrt} assume that backdoor triggers are robust and effective. 
When poisoned point clouds are superimposed by clean point clouds or subjected to varying degrees of destruction, 
they still consistently lead the infected 3D DNNs to predict the target label. 
This behavior allows for the identification of poisoned samples. 
Moreover, Spectral Signature~\cite{tran2018spectral} detects poisoned samples by analyzing the differences in latent space features between clean and poisoned samples. 
Besides, Grad-CAM~\cite{selvaraju2017grad} is also utilized to identify potential trigger regions within point clouds.

\begin{figure*}[t]
    \centering
    \includegraphics[width=0.9\textwidth]{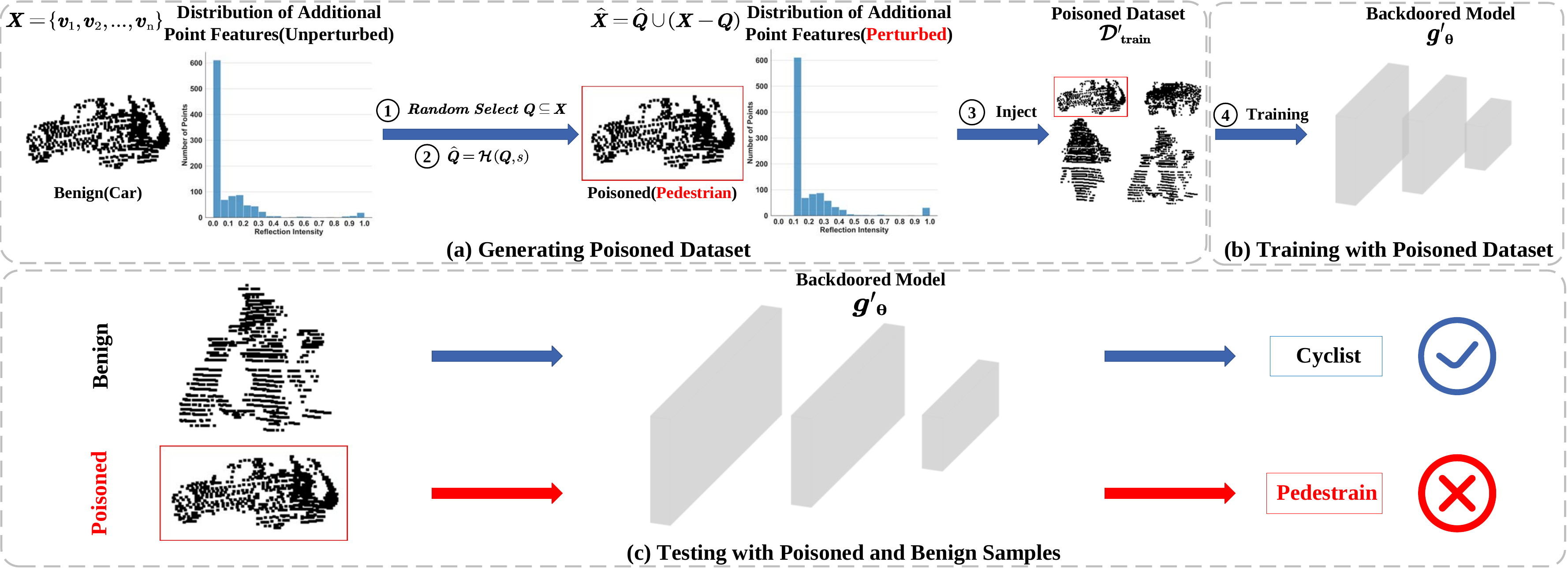}
    \caption{The attack pipeline of our proposed SRBA.}
	\label{fig:framework}
\end{figure*}

\section{Preliminaries}
\subsection{Threat Model}
We assume that attackers can access part of the training set and manipulate it to incorporate  malicious trigger and assign the target label. 
This assumption is consistent with previous works~\cite{li2021pointba,gao2023imperceptible} and is plausible in scenarios where the training process
is outsourced or users have collected some training data from openly available repositories polluted by attackers.  Attackers  are limited to  modifying the training data 
and cannot access or alter other parts involved in training process, such as model structure and training loss. Moreover, their goal is to make the backdoored models misclassify inputs with triggers as the target label,
while maintaining normal behavior on benign inputs.
\subsection{Problem Definition}
Consider a general 3D point cloud classification task. Let $\mathcal{D} _{\mathrm{train}}=\left\{ \left( X_{\mathrm{i}},y_{\mathrm{i}} \right) \right\} _{\mathrm{i}}^{N}$ be a training dataset consisting of  $N$ point clouds, where $y_{\mathrm{i}}\in \mathcal{Y} =\left\{ 1,2,...,K \right\} $
represents the category to which the point cloud $X_{\mathrm{i}}$ actually belongs, and $K$ represents the total number of categories.  Each point cloud $X_{\mathrm{i}}$ is a set of $n$ points $v_{\mathrm{i}}=\left( p_{\mathrm{i}},f_{\mathrm{i}} \right) $, where $p_{\mathrm{i}}=\left( x_{\mathrm{i}},y_{\mathrm{i}},z_{\mathrm{i}} \right) \in \mathbb{R} ^3$  and  $f_{\mathrm{i}}\in \mathbb{R} ^c$ represent the  
spatial position features and the additional point features of each point, respectively. Generally, the attackers will
select some samples from $\mathcal{D} _{\mathrm{train}}$ to craft the poisoned samples as $\mathcal{D} _{\mathrm{p}}=\left\{ \left( \hat{X}_{\mathrm{i}},y_{\mathrm{t}} \right) \right\} _{\mathrm{i}=1}^{M}$, where $y_{\mathrm{t}}$ is the target label. Additionally, $\eta =\frac{M}{N}$
denotes the poison rate, and $\mathcal{D} _{\mathrm{c}}=\left\{ \left( X_{\mathrm{i}},y_{\mathrm{i}} \right) \right\} _{\mathrm{i}=1}^{N-M}$ presents the rest of  benign sample set. The poisoned dataset ${\mathcal{D} ^{\prime}}_{\mathrm{train}}$ is ultimately composed of $\mathcal{D} _{\mathrm{c}}$ and $\mathcal{D} _{\mathrm{p}}$
as ${\mathcal{D} ^{\prime}}_{\mathrm{train}}=\mathcal{D} _{\mathrm{c}}\cup \mathcal{D} _{\mathrm{p}}$. 
The 3D DNN $g\left( X,\theta \right) \rightarrow y$ with parameters $\theta $ will be normally trained on  ${\mathcal{D} ^{\prime}}_{\mathrm{train}}$ via minimizing the following loss: 
{
	\small
	\begin{equation}
		\min_{\theta}\sum_{\left( X,y \right) \in \mathcal{D} _{\mathrm{train}}^{\prime}}{\mathcal{L} \left( g\left( \mathcal{P} \left( X \right) ,\theta \right) ,y \right)}
		\label{eq:Optimization Objective}
	\end{equation}}where $\mathcal{L} \left( \cdot ,\cdot \right) $ denotes the loss function and $\mathcal{P} \left( \cdot \right) $ represents the preprocessing operations applied to the training samples. 
Since these operations are commonly used for cleaning point clouds or enhancing the performance of 3D DNNs, 3D backdoor attacks that remain effective after the poisoned samples undergo preprocessing operations 
$\mathcal{P} \left( \cdot \right) $ are particularly threatening.

\section{Attack Design}
\subsection{Overview}
\subsubsection*{\textbf{Attack Intuition.}}
Unlike the existing 3D backdoor attacks typically manipulating the spatial position features to implant triggers, 
our attack focus on embedding the trigger through additional point features to achieve high stealthiness and resistance against various preprocessing
operations. In detail, our attack employs a uniform shift for additional point features of a subset of points within point clouds as the trigger. 
The intuition behind our attack stems from  the class-specific distributions observed in additional point features of point clouds, as illustrated in Fig.~\ref{fig:frequency}.
Hence, poisoned samples, where a uniform shift is employed to their additional point features, can  be distinguished from benign samples and be connected with target labels by 3D DNNs.
\subsubsection*{\textbf{Attack Pipeline.}}
Fig.~\ref{fig:framework} illustrates the pipeline of SRBA. 
SRBA consists of three main processes: (a) generating the poisoned dataset $\mathcal{D} _{\mathrm{train}}^{\prime}$, (b) training a 3D DNN with $\mathcal{D} _{\mathrm{train}}^{\prime}$,
and (c) testing with benign and poisoned point clouds. In detail, we first employ BO algorithm to identify an optimal trigger $s$. Subsequently, we  generate some poisoned point clouds using this trigger.
By merging all generated poisoned point clouds with the remaining benign ones, we create a malicious dataset $\mathcal{D} _{\mathrm{train}}^{\prime}$, which is publicly released for free use.
Upon using this malicious dataset to train their 3D DNNs, users inadvertently introduce our backdoor into their models.
When the infected 3D DNNs encounter the poisoned point clouds, our backdoor is triggered, causing the models to execute the intended malicious behavior.

Below we describe the details of our attack.
\subsection*{Searching the Optimal Trigger with BO}
Bayesian Optimization (BO)~\cite{shahriari2015taking} utilize a  probability surrogate model $\mathcal{M} $ (e.g., Gaussian Process)
to fit the prior distribution of an objective function from limited observations. The model $\mathcal{M} $ is updated with posterior knowledge gained from subsequent evaluations. 
By combining the model $\mathcal{M} $ with an acquisition function $\mathcal{A}$, like Expectation Improvement, 
BO intelligently selects new evaluation points, thereby reducing the number of evaluations required and approximating the optimal value of the objective function.
Hence, we utilize BO to automatically search for the optimal trigger $s$. Specifically,  
we train a surrogate backdoored model $g_{\mathrm{\theta}}^{\prime}$ for a few epochs using a poisoned dataset $\mathcal{D} _{\mathrm{train}}^{\prime}$,
as early training results can reliably reflect the final training performance~\cite{zoph2018learning}. 
We then assess the effectiveness of a trigger $s$
by evaluating the training loss on the poisoned point clouds and constrain the magnitude of $s$ using its L1 norm.
Formally, the objective function for employing BO to automatically dentify the optimal trigger can be expressed as follows:
{\small\begin{equation}
		O\left( s \right) =\sum_{i=1}^N{\mathcal{L} \left( g_{\mathrm{\theta}}^{\prime}\left( \hat{X}_{\mathrm{i}}^{s} \right) ,y_{\mathrm{t}} \right) +\lambda \left\| s \right\| _1}
	\label{eq:BO} 
	\end{equation}
}where $\hat{X}_{\mathrm{i}}^{s}$ denotes the poisoned point cloud generated with the trigger $s$ and $\lambda $ is a balancing weight. 
Furthermore, the detailed process for identifying the optimal trigger $s$ is outlined in Algorithm~\ref{al:BO}.

\subsection*{Generating Poisoned Point Clouds}
The key of our attack  lies in how to generate one poisoned point cloud $\hat{X}$.
To obtain $\hat{X}$, we first select points  
from the corresponding benign point cloud $X=\left\{ v_1,v_2,...,v_{\mathrm{n}} \right\} $ 
to obtain a subset $Q=\left\{ v_{\mathrm{i}_1},v_{\mathrm{i}_2},...,v_{\mathrm{i}_{\mathrm{w}}} \right\} \subseteq X$,
 where $w\in \left\{ 1,2,...,n \right\} $ is the size of $Q$. 
 In order to ensure selected points in $Q$ don't gather in one local region, we utilize Farthest Point Sampling (FPS) algorithm
 to choose points from $X$.
 After selecting points, 
 we employ a uniform shift $s\in \mathcal{S} \subset \mathbb{R} ^c$ (i.e., the trigger of SRBA) 
 on the additional point feature of each point in the $Q$ and obtain the $\hat{Q}$. This process is formulated as follows: 
{
	\small
	\begin{equation}
		\begin{split}
			\mathcal{H} \left( v_{\mathrm{j}},s \right) =\mathcal{H} \left( \left( p_{\mathrm{j}},f_{\mathrm{j}} \right) ,s \right) =\left( p_{\mathrm{j}},\mathcal{G} \left( f_{\mathrm{j}}+s \right) \right)  \\
			\hat{Q}=\mathcal{H} \left( Q,s \right) =\left\{ \mathcal{H} \left( v_{\mathrm{j}},s \right) \,\,|v_{\mathrm{j}}\in Q \right\} \\
		\end{split}
		\label{eq:trigger1}
	\end{equation}
}where $\mathcal{H} \left( \cdot ,\cdot \right) $ is the trigger implanting function and $\mathcal{G} \left( \cdot \right) $ is a data processing function (e.g., clipping  or unitization function), ensuring that values of the additional point features
remain within the normal range after a uniform shift. For instance, reflection intensity values should still stay within the normal range after the shift $s$ is employed. 
Ultimately, the poisoned sample $\hat{X}$ is obtained as $\hat{X}=\hat{Q}\cup \left( X-Q \right) $. 

\begin{algorithm}
	\small
	\caption{Search the optimal trigger with BO}
    \textbf{Input}: $\mathcal{M}$: probability surrogate model; $\mathcal{A} $: acquisition function; $T$: number of iteration; $Q$: number of initialized observation points \\
	\textbf{Output}: the optimal trigger $s$ 
	\begin{algorithmic}[1]
		\STATE Initialize observation set $\mathcal{F}$: $\mathcal{F} \gets \emptyset $
		\FOR{each observation point $i\gets 1$ to $Q$}
			\STATE Randomly initialize the trigger $s_{\mathrm{i}}$
			\STATE Calculate $O\left( s_{\mathrm{i}} \right) $ using Equation~\ref{eq:BO}
			\STATE $\mathcal{F}\gets \mathcal{F} \cup \left\{ \left( s_{\mathrm{i}},O\left( s_{\mathrm{i}} \right) \right) \right\} $
		\ENDFOR
		\FOR{$t\gets 1$ to $T$}
			\STATE Fit probability surrogate model $\mathcal{M}$: $\mathcal{M} \gets fit\left( \mathcal{F}  \right) $
			\STATE Select the next observation point $s_{\mathrm{t}}$ using  acquisition function $\mathcal{A} $: $s_{\mathrm{t}}=\mathop {\mathrm{arg}\,\min}_{s\in \mathcal{S}}\mathcal{A} \,\,(s|\mathcal{M} )$
			\STATE Calculate $O\left( s_{\mathrm{t}} \right) $ using Equation~\ref{eq:BO}
			\STATE $\mathcal{F}\gets \mathcal{F} \cup \left\{ \left( s_{\mathrm{t}},O\left( s_{\mathrm{t}} \right) \right) \right\} $
		\ENDFOR
		\STATE \textbf{return} $s_{\mathrm{best}}=\mathop {\mathrm{arg}\,\min}_{s\in \mathcal{F}}O(s)$
	\end{algorithmic}
	\label{al:BO}
\end{algorithm}

\begin{table*}[th]
	\centering
	\setlength{\tabcolsep}{1mm}
	\small
	\begin{tabular}{c|c|cc|cc|cc|cc|cc|cc} 
	\hline
	\multirow{2}{*}{Datasets}   & \multirow{2}{*}{Attacks} & \multicolumn{2}{c|}{PointNet}             & \multicolumn{2}{c|}{PointNet++}           & \multicolumn{2}{c|}{DGCNN}                & \multicolumn{2}{c|}{PointCNN}             & \multicolumn{2}{c|}{PCT}                  & \multicolumn{2}{c}{PT}                     \\ 
	\cline{3-14}
								&                          & \multicolumn{1}{c|}{ACC} & ASR            & \multicolumn{1}{c|}{ACC} & ASR            & \multicolumn{1}{c|}{ACC} & ASR            & \multicolumn{1}{c|}{ACC} & ASR            & \multicolumn{1}{c|}{ACC} & ASR            & \multicolumn{1}{c|}{ACC} & ASR             \\ 
	\hline
	\multirow{6}{*}{KITTI}      & No attack                & 86.70                    & –              & 86.72                    & –              & 86.54                    & –              & 85.30                    & –              & 86.10                    & –              & 86.83                    & –               \\
								& PointBA-I                & \textbf{86.80}           & \textbf{99.91} & 86.91                    & \textbf{99.99} & \underline{86.48}            & \textbf{99.99} & 85.10                    & 93.06          & 85.97                    & \textbf{99.93} & \textbf{86.69}           & \textbf{99.94}  \\
								& PointBA-O                & 85.74                    & 62.76          & \textbf{87.22}           & 91.92          & \textbf{86.67}           & 77.27          & \underline{85.39}            & 86.74          & \underline{86.01}            & 81.05          & \underline{86.39}            & 68.53           \\
								& IRBA                     & 86.15                    & 74.26          & 86.96                    & 89.54          & 86.22                    & 69.55          & \textbf{85.62}           & 86.17          & \textbf{86.42}           & 59.31          & 86.33                    & 66.19           \\
								& MirrorAttack             & 86.13                    & 86.05          & 86.24                    & \underline{99.38}          & 85.55                    & 97.74          & 84.31                    & \underline{97.72}  & 85.87                    & 99.50          & 85.66                    & 98.72           \\
								& SRBA (Ours)                     & \underline{86.50}        & \underline{98.68}  & \underline{87.04}            & 98.77          & 86.39                    & \underline{98.86}  & 85.04                    & \textbf{98.41} & 85.86                    & \underline{99.54}  & 86.14                    & \underline{98.85}   \\ 
	\hline
	\multirow{6}{*}{ModelNet10} & No attack                & 93.64                    & –              & 94.41                    & –              & 93.97                    & –              & 92.65                    & –              & 94.52                    & –              & 94.30                    & –               \\
								& PointBA-I                & 93.09                    & \textbf{100}   & \textbf{94.85}           & \textbf{100}   & 93.75                    & \textbf{100}   & \textbf{92.32}           & \underline{97.79}  & \textbf{94.08}           & \textbf{100}   & 93.86                    & \textbf{99.75}  \\
								& PointBA-O                & 92.98                    & 91.67          & \underline{94.30}            & 98.65          & \underline{93.97}            & 93.69          & 91.67                    & 93.63          & 93.86                    & 87.62          & \textbf{94.74}           & 99.51           \\
								& IRBA                     & \textbf{93.64}        & 92.83          & 94.30                    & 96.08          & 93.75                    & 90.44          & 91.97                    & 81.99          & 92.87                    & 75.86          & 93.64                    & 91.91           \\
								& MirrorAttack             & 92.96                    & 68.63          & 92.65                    & 99.26          & 93.20                    & \underline{99.26}  & 91.10                    & 97.67          & 92.43                    & 93.75          & 93.20                    & 98.90           \\
								& SRBA (Ours)              & \underline{93.31}         & \underline{99.02}  & 94.19                    & \underline{100}    & \textbf{94.08}           & 98.90          & \underline{92.32}            & \textbf{99.63} & \underline{94.08}            & \underline{100}    & \underline{93.86}            & \underline{99.75}   \\ 
	\hline
	\multirow{6}{*}{ModelNet40} & No attack                & 90.32                    & –              & 92.27                    & –              & 92.09                    & –              & 87.52                    & –              & 91.28                    & –              & 92.08                    & –               \\
								& PointBA-I                & \underline{90.60}        & \textbf{99.79} & 91.76                    & \textbf{99.87} & 91.45                    & \textbf{99.92} & 86.31                    & \underline{98.38}  & 90.26                    & \textbf{99.83} & \textbf{92.02}           & \textbf{99.79}  \\
								& PointBA-O                & 89.48                    & 88.68          & 91.84                    & 93.77          & \underline{91.64}            & 89.81          & \textbf{87.45}           & 73.55          & \textbf{91.04}           & 89.06          & 91.36                    & 94.02           \\
								& IRBA                     & 89.12                    & 78.56          & \textbf{92.08}           & 94.11          & \textbf{91.81}           & 78.51          & 85.54                    & 78.09          & 90.38                    & 69.65          & \underline{91.47}            & 85.25           \\
								& MirrorAttack             & 88.72                    & 64.84          & 91.42                    & \underline{99.12}  & 91.67                    & \underline{98.86}  & 84.35                    & \textbf{99.87} & \underline{91.03}            & 95.62          & 90.26                    & 97.77           \\
								& SRBA (Ours)                     & \textbf{90.72}           & \underline{94.40}  & \underline{92.02}            & 98.86          & 91.47                    & 97.85          & \underline{86.47}            & 96.84          & 90.71                    & \underline{96.80}  & 91.07                    & \underline{99.12}   \\
	\hline
	\end{tabular}
	\caption{Performance comparison across different backdoor attacks under all-to-one attack. The best and second-best value  are bolded ans underlined, respectively. 
	The ASR of SRBA ranks second only to the best attack in most cases, but it exhibits better stealthiness and robustness.} 
	\label{tb:attack results}
\end{table*}

\begin{table}[th]
	\centering
    \setlength{\tabcolsep}{1mm}
	\small
	\begin{tabular}{c|cc|cc|cc} 
	\hline
	\multirow{2}{*}{Models} & \multicolumn{2}{c|}{KITTI}       & \multicolumn{2}{c|}{ModelNet10}  & \multicolumn{2}{c}{ModelNet40}    \\ 
	\cline{2-7}
							& \multicolumn{1}{c|}{ACC} & ASR   & \multicolumn{1}{c|}{ACC} & ASR   & \multicolumn{1}{c|}{ACC} & ASR    \\ 
	\hline
	PointNet                & 86.25                    & 81.06 & 93.09                    & 78.95 & 89.32                    & 73.96  \\
	PointNet++              & 86.81                    & 82.64 & 94.52                    & 87.72 & 91.63                    & 75.79  \\
	DGCNN                   & 86.11                    & 72.54 & 93.75                    & 57.04 & 91.22                    & 38.84  \\
	PointCNN                & 85.28                    & 78.40 & 92.21                    & 81.32 & 86.61                    & 59.39  \\
	PCT                     & 85.27                    & 74.72 & 94.19                    & 72.08 & 90.59                    & 54.11  \\
	PT                      & 86.54                    & 81.33 & 93.42                    & 90.79 & 91.49                    & 83.54  \\
	\hline
	\end{tabular}
	\caption{Performance of SRBA under all-to-all attack.}
	\label{tb:all to all}
\end{table}

\section{Experiments}
\subsection{Experiment Setup}
To assess the effectiveness of our attack across diverse additional point features, 
we select ModelNet40, ModelNet10~\cite{wu20153d}, and KITTI~\cite{geiger2012we} as the benchmark datasets. 
ModelNet40 and ModelNet10 datasets include the face normal vector as an additional point feature, while KITTI incorporates reflection intensity.
We adopt widely used 3D DNNs as victim models, including PointNet~\cite{qi2017pointnet}, PointNet++~\cite{qi2017pointnet++}, PointCNN~\cite{li2018pointcnn}, DGCNN~\cite{wang2019dynamic}, PT~\cite{zhao2021point}, and PCT~\cite{guo2021pct}. 
Our attack is compared with four typical 3D backdoor attacks: PointBA-I~\cite{li2021pointba}, PointBA-O~\cite{li2021pointba}, IRBA~\cite{gao2023imperceptible}, and MirrorAttack~\cite{bian2024mirrorattack}. 
For all attacks, the poison rate $\eta$ is set to 0.05. We evaluate the effectiveness of 3D backdoor attacks using attack success rate (ASR) and assess the performance-preserving capability of the infected 3D DNNs through accuracy on clean samples (ACC). 
Additionally, we measure the stealthiness of these attacks using Wasserstein Distance (WD)~\cite{nguyen2021point}, 
which reveals the distributional differences between poisoned and clean point clouds. 
More details of the experiment setup are provided in the appendix.

\subsection{Effectiveness Evaluation}
\subsubsection*{\textbf{All-to-one Attack.}}
As illustrated in Table~\ref*{tb:attack results}, our attack 
consistently achieves an ASR of over 94$\%$ in all cases, slightly trailing behind the best attack. 
Despite not achieving the highest attack performances, our approach  offers distinct advantages such as robustness to preprocessing operations and greater imperceptibility,
which will be further analyzed  subsequently. 
Furthermore, compared to the ACC of the benign model, our attack results in a marginal decrease of no more than 1.5$\%$, demonstrating the models infected by our attack can maintain  
normal behavior on benign samples, and thus ensuring stealthiness.

\subsubsection*{\textbf{All-to-all Attack.}}
In contrast to all-to-one attacks, all-to-all attacks assign
different target labels to different poisoned samples.
We also evaluate the performance of our attack under the all-to-all attack scenario following~\cite{gu2019badnets}. 
The results presented in Table~\ref{tb:all to all} indicate that
SRBA continues to achieve reliable attack performances.

\begin{table}[t]
	\centering
	\small
	\begin{tabular}{c|c|c|c} 
	\hline
	Methods      & ModelNet10    & ModelNet40    & KITTI          \\ 
	\hline
	PointBA-I    & 0.94          & 1.03          & 5.05           \\
	PointBA-O    & \textbf{0.19} & \textbf{0.24} & 1.17           \\
	IRBA         & 1.37          & 1.40          & 6.03           \\
	MirrorAttack & 0.47          & 0.59          & 2.67           \\
	\hline
	SRBA*        & 0.73          & 0.73          & \textbf{0.63}  \\
	SRBA (Ours)   & 2.92          & 2.89          & 2.53           \\
	\hline
	\end{tabular}
	\caption{Wasserstein distance between benign and poisoned point clouds. The best value is bolded. * indicates adopting
	a smaller trigger.}
	\label{tb:st}
\end{table}
\subsection{Stealthiness Evaluation}
As demonstrated in Table~\ref{tb:st}, when adopting a smaller trigger, SRBA shows competitive performance on the WD metric, 
and achieves the optimal result on the KITTI dataset. The trade-off for using a smaller trigger is minimal, with the ASR only slightly decreasing to 93.75\%, 92.07\%, and 98.55\% on the ModelNet10, ModelNet40, and KITTI datasets, respectively.
Furthermore, since SRBA embeds its trigger within the additional point features of point clouds without altering their geometric information, 
SRBA maintains visual stealthiness even when encountering more complex point clouds, as illustrated in Fig.~\ref{fig:compare}.
Besides, more examples similar to Fig.~\ref{fig:compare} are provided in the appendix.

\begin{table*}[ht]
    \centering
    \setlength{\tabcolsep}{1mm}
	\small
    \begin{tabular}{ccccccc|cc|cc|cc|cc|cc} 
    \hline
    \multicolumn{7}{c|}{Preprocessing Operations}                                                                                                                           & \multicolumn{10}{c}{Method}                                                                                                                                                                      \\ 
    \hline
    \multirow{2}{*}{SOR} & \multirow{2}{*}{Rotation} & \multirow{2}{*}{Rotation-3D} & \multirow{2}{*}{Scaling} & \multirow{2}{*}{Shift} & \multirow{2}{*}{Dropout} & \multirow{2}{*}{Jitter} & \multicolumn{2}{c|}{PointBA-I}            & \multicolumn{2}{c|}{PointBA-O}   & \multicolumn{2}{c|}{IRBA}        & \multicolumn{2}{c|}{MirrorAttack} & \multicolumn{2}{c}{SRBA (Ours)}                   \\ 
    \cline{8-17}
                         &                    &                     &                          &                        &                          &                         & \multicolumn{1}{c|}{ACC} & ASR            & \multicolumn{1}{c|}{ACC} & ASR   & \multicolumn{1}{c|}{ACC} & ASR   & \multicolumn{1}{c|}{ACC} & ASR    & \multicolumn{1}{c|}{ACC} & ASR             \\ 
    \hline
    \ding{55}            & \ding{55}          & \ding{55}           & \ding{55}                & \ding{55}              & \ding{55}                & \ding{55}               & \textbf{86.80}           & \textbf{99.91} & 85.74                    & 62.76 & 86.15                    & 74.26 & 86.13                    & 86.05  & 86.50                    & 98.68           \\
    \checkmark           & \ding{55}          & \ding{55}           & \ding{55}                & \ding{55}              & \ding{55}                & \ding{55}               & 85.87                    & 0.03           & 84.51                    & 61.65 & 85.82                    & 70.81 & 83.48                    & 75.36  & \textbf{86.16}           & \textbf{99.16}  \\
    \checkmark           & \checkmark         & \ding{55}           & \ding{55}                & \ding{55}              & \ding{55}                & \ding{55}               & 85.57                    & 0.37           & 85.53                    & 1.00  & 85.31                    & 70.69 & 84.29                    & 75.90  & \textbf{85.70}           & \textbf{99.48}  \\
    \checkmark           & \checkmark         & \checkmark          & \ding{55}                & \ding{55}              & \ding{55}                & \ding{55}               & 79.79                    & 4.26           & 79.95                    & 1.31  & \textbf{80.42}           & 50.31 & 76.04                    & 11.07  & 80.10                    & \textbf{99.76}  \\
    \checkmark           & \checkmark         & \checkmark          & \checkmark               & \ding{55}              & \ding{55}                & \ding{55}               & 79.86                    & 16.85          & 79.53                    & 0.92  & 80.27                    & 50.78 & 75.25                    & 12.27  & \textbf{80.72}           & \textbf{99.51}  \\
    \checkmark           & \checkmark         & \checkmark          & \checkmark               & \checkmark             & \ding{55}                & \ding{55}               & 79.66                    & 5.60           & 79.68                    & 1.22  & 79.94                    & 46.60 & 78.47                    & 12.86  & \textbf{80.65}           & \textbf{99.75}  \\
    \checkmark           & \checkmark         & \checkmark          & \checkmark               & \checkmark             & \checkmark               & \ding{55}               & 79.63                    & 5.19           & 79.58                    & 0.97  & 79.94                    & 39.70 & 75.65                    & 10.72  & \textbf{80.51}           & \textbf{99.51}  \\
    \checkmark           & \checkmark         & \checkmark          & \checkmark               & \checkmark             & \checkmark               & \checkmark              & 79.57                    & 4.47           & 79.59                    & 1.52  & 80.36                    & 21.56 & 74.23                    & 3.82   & \textbf{80.41}           & \textbf{99.57}  \\
    \hline
    \end{tabular}
	\caption{Resistance results against preprocessing-based defenses (KITTI).
    The best value is bolded. "\checkmark" indicates including that operation during training
    and "\ding{55}" indicates not.}
	\label{tb:aug defense}
\end{table*}

\subsection{Robustness Evaluation}
In this subsection, we use the ModelNet10 and KITTI as the default datasets, with PointNet as the default victim model.
\subsubsection{\textbf{Resistance to Preprocessing-based Defenses.}}
Common preprocessing operations adopted in this work contain Statistical Outlier Removal (SOR), Rotation, Rotation-3D, Scaling, Shift, Dropout, and
Jitter. SOR eliminates points in a point cloud that are significantly more distant from their neighbors than the average
distance. We set the hyperparameters $k$ (i.e., number of nearest neighbors) and standard deviation $\delta $ to 30 and 2, respectively. Rotation involves  randomly rotating
point clouds up to $20^{\circ}$ around the x-axis, while Rotation-3D perform random rotation up to $360^{\circ}$ around all axes.
The scaling factor and shift distance are randomly sampled from uniform distribution $\mathcal{U} \left( 0.5,1.5 \right) $ and $\mathcal{U} \left( \left[ -0.1,0.1 \right] ^3 \right) $, respectively. 
Dropout randomly removes  0\% to 50\% of points in point clouds, and Jitter adds point-wise Gaussian noise $\mathcal{N} \left( 0,0.02^2 \right) $ to point clouds.
Here, we present the preprocessing-based defense results on KITTI, with  the results on ModelNet10 in the appendix.
As presented in Table~\ref*{tb:aug defense}, 
our attack achieves exceptionally high ASRs when confronted with all preprocessing operations simultaneously. However, the ASRs of other comparative attacks decline significantly. Since preprocessing
operations target the spatial position features of point clouds, our attack operates on the additional point features, making preprocessing-based defenses ineffective against our attack.

\begin{figure}[t]
    \centering
    \includegraphics[width=0.45\textwidth]{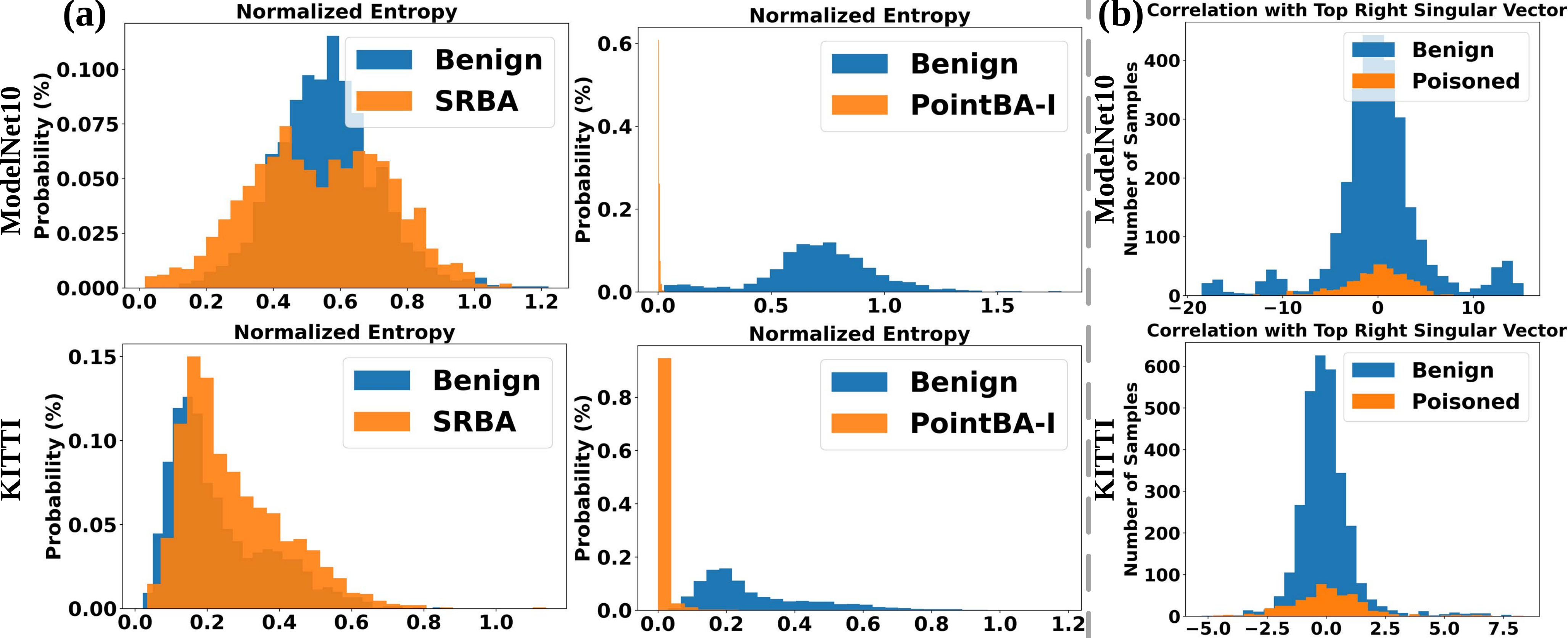}
    \caption{(a) Resistance results against STRIP. (b) Resistance results against Spectral Signature.}
    \label{fig:strip ssd}
\end{figure}

\begin{figure}[t]
    \centering
    \includegraphics[width=0.45\textwidth]{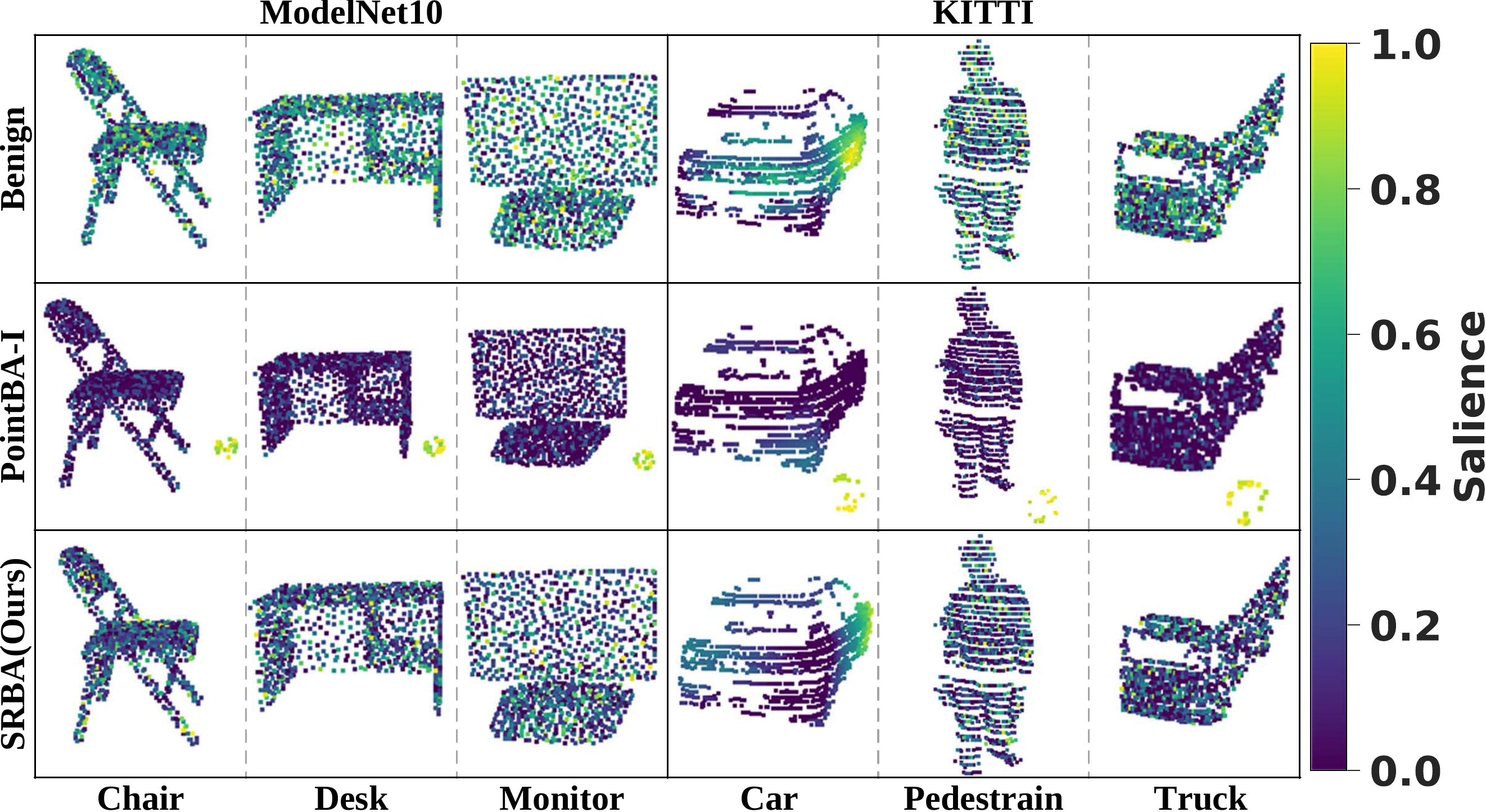}
    \caption{Resistance results against 3D Grad-CAM.}
    \label{fig:grad cam}
\end{figure}

\begin{figure}[t]
    \centering
    \includegraphics[width=0.45\textwidth]{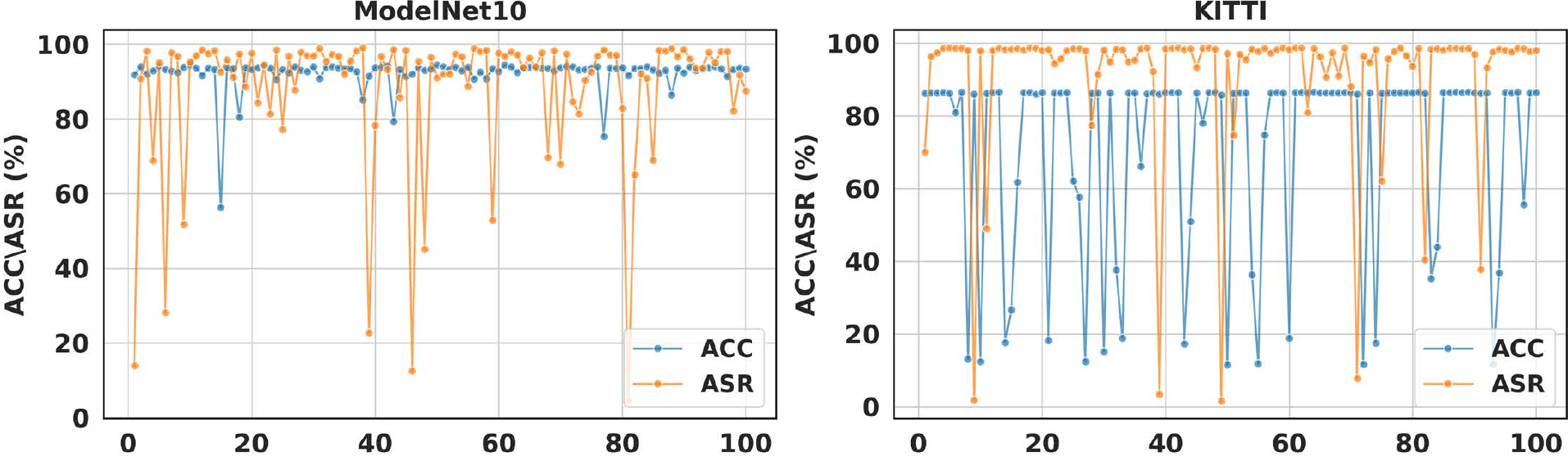}
    \caption{Defense effect of adaptive defense during inference.}
    \label{fig:test}
\end{figure}
\begin{table}[t]
	\centering
	\small
	\begin{tabular}{c|cc|cc} 
	\hline
	\multirow{2}{*}{w} & \multicolumn{2}{c|}{ModelNet10}  & \multicolumn{2}{c}{KITTI}         \\ 
	\cline{2-5}
					   & \multicolumn{1}{c|}{ACC} & ASR   & \multicolumn{1}{c|}{ACC} & ASR    \\ 
	\hline
	200                & 92.76                    & 73.35 & 85.99                    & 83.19  \\
	400                & 92.21                    & 76.90 & 86.25                    & 83.82  \\
	600                & 93.42                    & 81.68 & 86.36                    & 83.63  \\
	800                & 93.09                    & 81.80 & 85.95                    & 84.43  \\
	1024               & 92.32                    & 39.28 & 86.66                    & 32.86  \\
	\hline
	\end{tabular}
	\caption{Defense effect of adaptive defense during training.}
	\label{tb:test}
\end{table}

\subsubsection{\textbf{Resistance to STRIP.}}
STRIP~\cite{gao2019strip} detects poisoned samples if the predictions of these samples superimposed with different benign samples exhibit consistently low entropy. Thus, the effectiveness of STRIP can be assessed by comparing the disparity in entropy distribution between benign and poisoned samples.
Results in Fig.~\ref{fig:strip ssd} (a) indicate that, for SRBA, the distributions of these two types of samples are nearly identical, thus STRIP is unable to differentiate its poisoned samples. Conversely, STRIP effectively differentiates poisoned samples from PointBA-I. 
This distinction arises because the superimposing operation with benign samples disrupts SRBA's trigger, but not PointBA-I's trigger. Consequently, the predictions of a poisoned sample superimposed with various benign samples will change randomly rather than consistently aligning with the intended target labels,
similar to a benign sample superimposed with various benign samples.

\subsubsection*{\textbf{Resistance to Spectral Signature.}}
Spectral Signature~\cite{tran2018spectral} detects potential poisoned samples using latent space features.   
To evaluate the effectiveness of Spectral Signature, we randomly select 3,000 benign and 500 poisoned samples from each dataset and plot the histograms of correlation scores for both sets of samples following~\cite{tran2018spectral}. 
Fig.~\ref{fig:strip ssd} (b) show that there is no clear distinction in correlation scores between benign and poisoned samples.
Therefore, this method struggles to detect poisoned samples effectively in our attack scenario, thus failing to defend against our attack.

\subsubsection{\textbf{Resistance to 3D Grad-CAM.}}
Grad-CAM~\cite{selvaraju2017grad} provides the visual interpretation of  the model's behavior during inference and can be used to detect potential trigger regions. 
It identifys the backdoor trigger by finding a small, unusual region that significantly determines the prediction results. 
We extend Grad-CAM to point clouds following ~\cite{bian2024mirrorattack}.
Fig.~\ref{fig:grad cam} demonstrates that 3D Grad-CAM can effectively identify the trigger region for PointBA-I.  
However, it fails to locate the trigger region in poisoned samples of SRBA, 
as the distributions of salient points in their heatmaps appear sparse and scattered rather than concentrated in specific areas, 
similar to benign samples.
This is because SRBA employs a uniform shift on the additional point features of scattered points across the entire point clouds, 
while PointBA-I affects only localized regions.

\subsubsection*{\textbf{Resistance to Adaptive Defense.}}
We consider an adaptive defense against SRBA where the defender, 
aware of the presence of SRBA but lacking precise knowledge about its parameters like  the shift $s$ and size $w$ of $Q$,
perturbs the distributions of additional point features within each point cloud during the training or inference phase.
Specifically, this perturbation is achieved simply by uniformly adding Gaussian noise  to 
the additional point features of all points within each point cloud before sending it to the victim model. Fig.~\ref{fig:test} presents  results from 100 
repeated experiments during inference, revealing large fluctuations in ASR and ACC due to the randomness of Gaussian noise.
Furthermore, we observe that this defense is effective only when the added Gaussian noise approximately counteracts the  distribution perturbation caused by SRBA in additional point features.
Table.~\ref{tb:test} details  this defense's performance against SRBA with different $w$ values during training, consistently showing SRBA's ability to sustain its attack effectiveness.
Despite the addition of Gaussian noise, the distributions of additional point features in poisoned and benign samples remain distinct enough for the victim model to correctly associate poisoned samples with the target label during training.
In summary, this adaptive defense proves ineffective against SRBA due to the lack of precise information about its parameters.

\begin{figure}[t]
	\centering
    \includegraphics[width=0.45\textwidth]{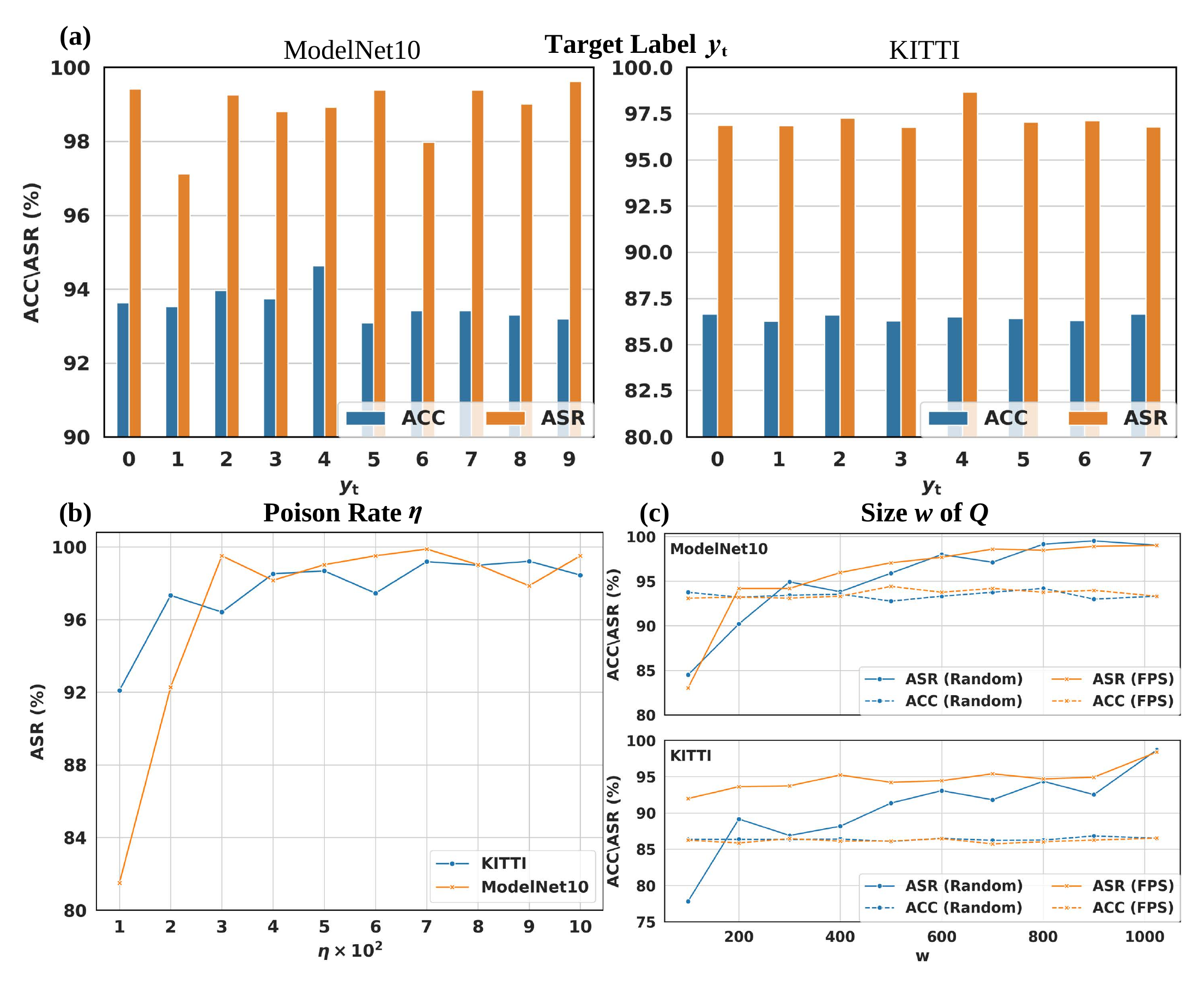}
    \caption{Effect of the different target label $y_{\mathrm{t}}$, poison rate  $\eta $  and size $w$ of $Q$.}
    \label{fig:poison rate part}
\end{figure}
\begin{figure}[th]
	\centering
    \includegraphics[width=0.45\textwidth]{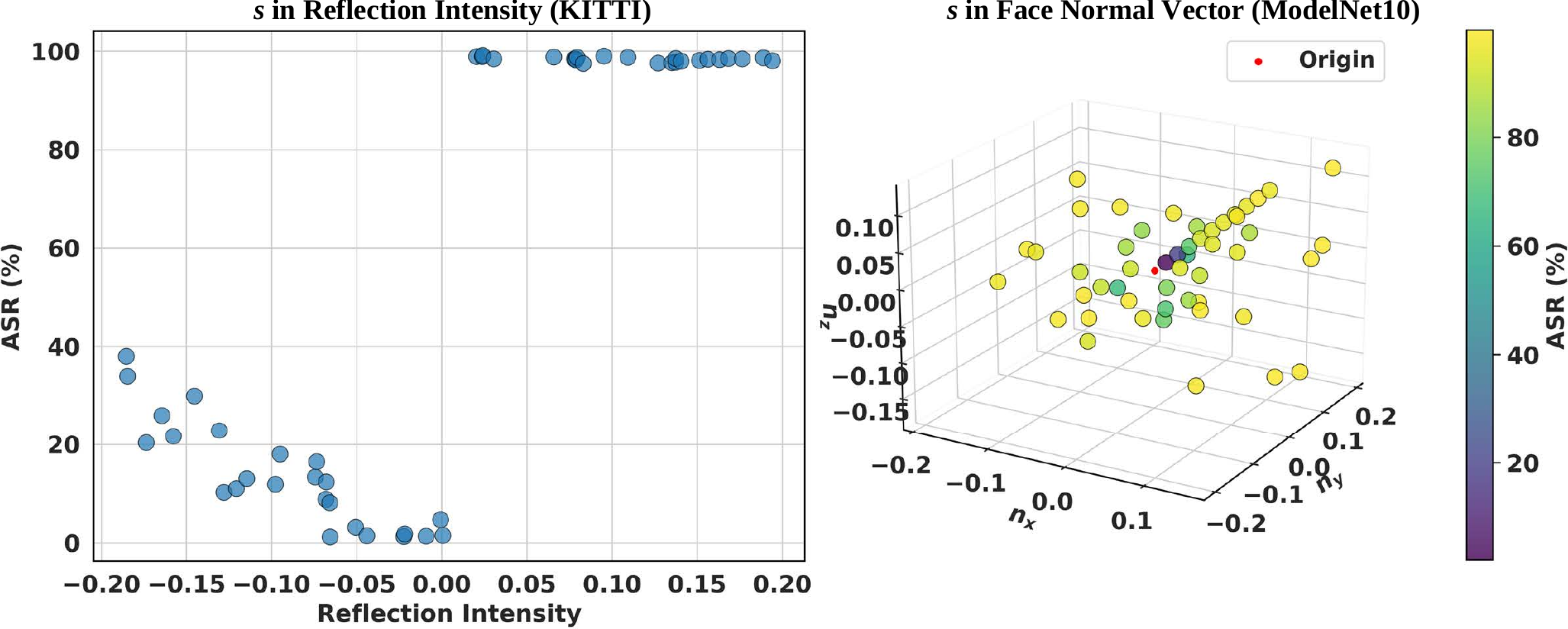}
    \caption{Effect of the different shift $s$.}
    \label{fig:shift}
\end{figure}

\subsection{Ablation Studies}
For simplicity, we conduct all the experiments with PointNet on ModelNet10  and KITTI datasets in this subsection.
\subsubsection*{\textbf{Effect of the Target Label $y_{\mathrm{t}}$ and Poison Rate $\eta $.}}
We investigate the effects of different target labels and poison rates on our attack in Fig.~\ref{fig:poison rate part} (a) and (b), respectively.
Fig.~\ref{fig:poison rate part} (a) demonstrates that varying target labels only slightly affects  both ASR and ACC. 
Furthermore, Fig.~\ref{fig:poison rate part} (b) shows that the ASR increases with increasing poison rates.
Notably, when the poison rate reaches around 2\%, our method can demonstrate significant attack performance.

\subsubsection*{\textbf{Effect of the Size $w$ of the Subset $Q$.}}
Here, we present the results of our attack under the varying size $w$ of $Q$ in Fig.~\ref{fig:poison rate part} (c). 
As the size $w$ increases, the ASR of our attack gradually rises while having minimal impact on the ACC. Remarkably, 
even when we employ a uniform shift on the additional point features of  only 100 points in a point cloud, we still achieve an ASR of around $80\%$.
This is because applying a uniform shift on just a hundred points in the point cloud also changes the distribution of its additional point features enough to differentiate it from the original distribution. 
Although this difference is subtle, the 3D DNNs can still learn to associate this specific distribution of poisoned point clouds with the target label during training, thereby embedding our backdoor. 
Furthermore, Fig.~\ref{fig:poison rate part} (c)  also illustrates that the impact of random selection versus FPS for choosing subset 
$Q$ on attack performance is negligible. However, FPS performs slightly better on the KITTI when $w$ is small.

\subsubsection*{\textbf{Effect of the Shift $s$.}}
To investigate the effect of different $s$ (i.e., the trigger of SRBA) on attack performance, 
we conduct 50 repeated experiments, randomly selecting each shift $s$. 
Fig.~\ref{fig:shift} illustrates the ASR of SRBA under the various $s$, demonstrating the universality of the shift. This
indicates that most values are practical and can be manually selected according to  attackers' requirements. 
Nevertheless, to further enhance the practicality and comprehensiveness of our attack strategy,
we still introduce BO to find the appropriate shift.

\section{Conclusions}
\label{con}
In this paper, we propose a novel poisoning-based backdoor attack against 3D point clouds, named SRBA, 
which utilizes a uniform shift in the additional point features of partial points within each point cloud as the trigger. 
The poisoned samples created by our method are stealthy and can bypass defender inspections. Extensive experiment results demonstrates
that SRBA achieves the superior attack performance while maintaining the robustness against preprocessing-based defenses and other mainstream backdoor defenses.

\bibliography{aaai25}

\begin{thebibliography}{35}
\providecommand{\natexlab}[1]{#1}

\bibitem[{Bian, Tian, and Liu(2024)}]{bian2024mirrorattack}
Bian, Y.; Tian, S.; and Liu, X. 2024.
\newblock MirrorAttack: Backdoor Attack on 3D Point Cloud with a Distorting Mirror.
\newblock \emph{arXiv preprint arXiv:2403.05847}.

\bibitem[{Fan et~al.(2022)Fan, He, Guo, Tang, Hong, and Li}]{fan2022careful}
Fan, L.; He, F.; Guo, Q.; Tang, W.; Hong, X.; and Li, B. 2022.
\newblock Be careful with rotation: A uniform backdoor pattern for 3D shape.
\newblock \emph{arXiv preprint arXiv:2211.16192}.

\bibitem[{Fan et~al.(2024)Fan, He, Si, Tang, and Li}]{fan2024invisible}
Fan, L.; He, F.; Si, T.; Tang, W.; and Li, B. 2024.
\newblock Invisible Backdoor Attack against 3D Point Cloud Classifier in Graph Spectral Domain.
\newblock In \emph{Proceedings of the AAAI Conference on Artificial Intelligence}, volume~38, 21072--21080.

\bibitem[{Feng et~al.(2024)Feng, Qian, Zhang, and Li}]{feng2024stealthy}
Feng, L.; Qian, Z.; Zhang, X.; and Li, S. 2024.
\newblock Stealthy Backdoor Attacks On Deep Point Cloud Recognization Networks.
\newblock \emph{The Computer Journal}, 67(5): 1879--1891.

\bibitem[{Gao et~al.(2023)Gao, Bai, Wu, Ya, and Xia}]{gao2023imperceptible}
Gao, K.; Bai, J.; Wu, B.; Ya, M.; and Xia, S.-T. 2023.
\newblock Imperceptible and robust backdoor attack in 3d point cloud.
\newblock \emph{IEEE Transactions on Information Forensics and Security}, 19: 1267--1282.

\bibitem[{Gao et~al.(2019)Gao, Xu, Wang, Chen, Ranasinghe, and Nepal}]{gao2019strip}
Gao, Y.; Xu, C.; Wang, D.; Chen, S.; Ranasinghe, D.~C.; and Nepal, S. 2019.
\newblock Strip: A defence against trojan attacks on deep neural networks.
\newblock In \emph{Proceedings of the 35th annual computer security applications conference}, 113--125.

\bibitem[{Geiger, Lenz, and Urtasun(2012)}]{geiger2012we}
Geiger, A.; Lenz, P.; and Urtasun, R. 2012.
\newblock Are we ready for autonomous driving? the kitti vision benchmark suite.
\newblock In \emph{2012 IEEE conference on computer vision and pattern recognition}, 3354--3361. IEEE.

\bibitem[{Gu, Dolan-Gavitt, and Garg(2017)}]{gu2017badnets}
Gu, T.; Dolan-Gavitt, B.; and Garg, S. 2017.
\newblock Badnets: Identifying vulnerabilities in the machine learning model supply chain.
\newblock \emph{arXiv preprint arXiv:1708.06733}.

\bibitem[{Gu et~al.(2019)Gu, Liu, Dolan-Gavitt, and Garg}]{gu2019badnets}
Gu, T.; Liu, K.; Dolan-Gavitt, B.; and Garg, S. 2019.
\newblock Badnets: Evaluating backdooring attacks on deep neural networks.
\newblock \emph{IEEE Access}, 7: 47230--47244.

\bibitem[{Guo et~al.(2021)Guo, Cai, Liu, Mu, Martin, and Hu}]{guo2021pct}
Guo, M.-H.; Cai, J.-X.; Liu, Z.-N.; Mu, T.-J.; Martin, R.~R.; and Hu, S.-M. 2021.
\newblock Pct: Point cloud transformer.
\newblock \emph{Computational Visual Media}, 7: 187--199.

\bibitem[{Hammoud and Ghanem(2021)}]{hammoud2021check}
Hammoud, H. A. A.~K.; and Ghanem, B. 2021.
\newblock Check your other door! Creating backdoor attacks in the frequency domain.
\newblock \emph{arXiv preprint arXiv:2109.05507}.

\bibitem[{Hu et~al.(2023)Hu, Liu, Li, Zhang, Liu, Wang, Zhang, and Hou}]{hu2023pointcrt}
Hu, S.; Liu, W.; Li, M.; Zhang, Y.; Liu, X.; Wang, X.; Zhang, L.~Y.; and Hou, J. 2023.
\newblock Pointcrt: Detecting backdoor in 3d point cloud via corruption robustness.
\newblock In \emph{Proceedings of the 31st ACM International Conference on Multimedia}, 666--675.

\bibitem[{Jiang et~al.(2023)Jiang, Li, Xu, and Zhang}]{jiang2023color}
Jiang, W.; Li, H.; Xu, G.; and Zhang, T. 2023.
\newblock Color backdoor: A robust poisoning attack in color space.
\newblock In \emph{Proceedings of the IEEE/CVF Conference on Computer Vision and Pattern Recognition}, 8133--8142.

\bibitem[{Kim et~al.(2021)Kim, Lee, Hwang, Lee, Hwang, and Kim}]{kim2021point}
Kim, S.; Lee, S.; Hwang, D.; Lee, J.; Hwang, S.~J.; and Kim, H.~J. 2021.
\newblock Point cloud augmentation with weighted local transformations.
\newblock In \emph{Proceedings of the IEEE/CVF international conference on computer vision}, 548--557.

\bibitem[{Li et~al.(2020)Li, Xue, Zhao, Zhu, and Zhang}]{li2020invisible}
Li, S.; Xue, M.; Zhao, B. Z.~H.; Zhu, H.; and Zhang, X. 2020.
\newblock Invisible backdoor attacks on deep neural networks via steganography and regularization.
\newblock \emph{IEEE Transactions on Dependable and Secure Computing}, 18(5): 2088--2105.

\bibitem[{Li et~al.(2021)Li, Chen, Zhao, Tong, Zhao, Lim, and Zhou}]{li2021pointba}
Li, X.; Chen, Z.; Zhao, Y.; Tong, Z.; Zhao, Y.; Lim, A.; and Zhou, J.~T. 2021.
\newblock Pointba: Towards backdoor attacks in 3d point cloud.
\newblock In \emph{Proceedings of the IEEE/CVF international conference on computer vision}, 16492--16501.

\bibitem[{Li et~al.(2018)Li, Bu, Sun, Wu, Di, and Chen}]{li2018pointcnn}
Li, Y.; Bu, R.; Sun, M.; Wu, W.; Di, X.; and Chen, B. 2018.
\newblock Pointcnn: Convolution on x-transformed points.
\newblock \emph{Advances in neural information processing systems}, 31.

\bibitem[{Nguyen et~al.(2021)Nguyen, Pham, Le, Pham, Ho, and Hua}]{nguyen2021point}
Nguyen, T.; Pham, Q.-H.; Le, T.; Pham, T.; Ho, N.; and Hua, B.-S. 2021.
\newblock Point-set distances for learning representations of 3d point clouds.
\newblock In \emph{Proceedings of the IEEE/CVF international conference on computer vision}, 10478--10487.

\bibitem[{Qi et~al.(2017{\natexlab{a}})Qi, Su, Mo, and Guibas}]{qi2017pointnet}
Qi, C.~R.; Su, H.; Mo, K.; and Guibas, L.~J. 2017{\natexlab{a}}.
\newblock Pointnet: Deep learning on point sets for 3d classification and segmentation.
\newblock In \emph{Proceedings of the IEEE conference on computer vision and pattern recognition}, 652--660.

\bibitem[{Qi et~al.(2017{\natexlab{b}})Qi, Yi, Su, and Guibas}]{qi2017pointnet++}
Qi, C.~R.; Yi, L.; Su, H.; and Guibas, L.~J. 2017{\natexlab{b}}.
\newblock Pointnet++: Deep hierarchical feature learning on point sets in a metric space.
\newblock \emph{Advances in neural information processing systems}, 30.

\bibitem[{Selvaraju et~al.(2017)Selvaraju, Cogswell, Das, Vedantam, Parikh, and Batra}]{selvaraju2017grad}
Selvaraju, R.~R.; Cogswell, M.; Das, A.; Vedantam, R.; Parikh, D.; and Batra, D. 2017.
\newblock Grad-cam: Visual explanations from deep networks via gradient-based localization.
\newblock In \emph{Proceedings of the IEEE international conference on computer vision}, 618--626.

\bibitem[{Shahriari et~al.(2015)Shahriari, Swersky, Wang, Adams, and De~Freitas}]{shahriari2015taking}
Shahriari, B.; Swersky, K.; Wang, Z.; Adams, R.~P.; and De~Freitas, N. 2015.
\newblock Taking the human out of the loop: A review of Bayesian optimization.
\newblock \emph{Proceedings of the IEEE}, 104(1): 148--175.

\bibitem[{Tran, Li, and Madry(2018)}]{tran2018spectral}
Tran, B.; Li, J.; and Madry, A. 2018.
\newblock Spectral signatures in backdoor attacks.
\newblock \emph{Advances in neural information processing systems}, 31.

\bibitem[{Wang et~al.(2021)Wang, Yao, Xu, An, Tong, and Wang}]{wang2021backdoor}
Wang, T.; Yao, Y.; Xu, F.; An, S.; Tong, H.; and Wang, T. 2021.
\newblock Backdoor attack through frequency domain.
\newblock \emph{arXiv preprint arXiv:2111.10991}.

\bibitem[{Wang et~al.(2019)Wang, Sun, Liu, Sarma, Bronstein, and Solomon}]{wang2019dynamic}
Wang, Y.; Sun, Y.; Liu, Z.; Sarma, S.~E.; Bronstein, M.~M.; and Solomon, J.~M. 2019.
\newblock Dynamic graph cnn for learning on point clouds.
\newblock \emph{ACM Transactions on Graphics (tog)}, 38(5): 1--12.

\bibitem[{Wu et~al.(2015)Wu, Song, Khosla, Yu, Zhang, Tang, and Xiao}]{wu20153d}
Wu, Z.; Song, S.; Khosla, A.; Yu, F.; Zhang, L.; Tang, X.; and Xiao, J. 2015.
\newblock 3d shapenets: A deep representation for volumetric shapes.
\newblock In \emph{Proceedings of the IEEE conference on computer vision and pattern recognition}, 1912--1920.

\bibitem[{Xiang et~al.(2021)Xiang, Miller, Chen, Li, and Kesidis}]{xiang2021backdoor}
Xiang, Z.; Miller, D.~J.; Chen, S.; Li, X.; and Kesidis, G. 2021.
\newblock A backdoor attack against 3d point cloud classifiers.
\newblock In \emph{Proceedings of the IEEE/CVF international conference on computer vision}, 7597--7607.

\bibitem[{Yang et~al.(2018)Yang, Feng, Shen, and Tian}]{yang2018foldingnet}
Yang, Y.; Feng, C.; Shen, Y.; and Tian, D. 2018.
\newblock Foldingnet: Point cloud auto-encoder via deep grid deformation.
\newblock In \emph{Proceedings of the IEEE conference on computer vision and pattern recognition}, 206--215.

\bibitem[{Yang et~al.(2020)Yang, Sun, Liu, and Jia}]{yang20203dssd}
Yang, Z.; Sun, Y.; Liu, S.; and Jia, J. 2020.
\newblock 3dssd: Point-based 3d single stage object detector.
\newblock In \emph{Proceedings of the IEEE/CVF conference on computer vision and pattern recognition}, 11040--11048.

\bibitem[{Yin, Zhou, and Krahenbuhl(2021)}]{yin2021center}
Yin, T.; Zhou, X.; and Krahenbuhl, P. 2021.
\newblock Center-based 3d object detection and tracking.
\newblock In \emph{Proceedings of the IEEE/CVF conference on computer vision and pattern recognition}, 11784--11793.

\bibitem[{Zhao et~al.(2021)Zhao, Jiang, Jia, Torr, and Koltun}]{zhao2021point}
Zhao, H.; Jiang, L.; Jia, J.; Torr, P.~H.; and Koltun, V. 2021.
\newblock Point transformer.
\newblock In \emph{Proceedings of the IEEE/CVF international conference on computer vision}, 16259--16268.

\bibitem[{Zhong et~al.(2020)Zhong, Liao, Squicciarini, Zhu, and Miller}]{zhong2020backdoor}
Zhong, H.; Liao, C.; Squicciarini, A.~C.; Zhu, S.; and Miller, D. 2020.
\newblock Backdoor embedding in convolutional neural network models via invisible perturbation.
\newblock In \emph{Proceedings of the Tenth ACM Conference on Data and Application Security and Privacy}, 97--108.

\bibitem[{Zhou et~al.(2019)Zhou, Chen, Zhang, Fang, Zhou, and Yu}]{zhou2019dup}
Zhou, H.; Chen, K.; Zhang, W.; Fang, H.; Zhou, W.; and Yu, N. 2019.
\newblock Dup-net: Denoiser and upsampler network for 3d adversarial point clouds defense.
\newblock In \emph{Proceedings of the IEEE/CVF international conference on computer vision}, 1961--1970.

\bibitem[{Zhou and Tuzel(2018)}]{zhou2018voxelnet}
Zhou, Y.; and Tuzel, O. 2018.
\newblock Voxelnet: End-to-end learning for point cloud based 3d object detection.
\newblock In \emph{Proceedings of the IEEE conference on computer vision and pattern recognition}, 4490--4499.

\bibitem[{Zoph et~al.(2018)Zoph, Vasudevan, Shlens, and Le}]{zoph2018learning}
Zoph, B.; Vasudevan, V.; Shlens, J.; and Le, Q.~V. 2018.
\newblock Learning transferable architectures for scalable image recognition.
\newblock In \emph{Proceedings of the IEEE conference on computer vision and pattern recognition}, 8697--8710.

\end{thebibliography}

\end{document}